# *Correlated Anomaly Detection from Large Streaming Data*

Zheng Chen[1*], Xinli Yu[2*], Yuan Ling[3], Bo Song[1], Wei Quan[1], Xiaohua Hu[1], Erjia Yan[1]

[1]College of Computing & Informatics, Drexel University
[2]Department of Mathematics, Temple University
[3]Alexa AI, Amazon Inc.

*ABSTRACT* — Correlated anomaly detection (CAD) from streaming data is a type of group anomaly detection and an essential task in useful real-time data mining applications like botnet detection, financial event detection, industrial process monitor, etc. The primary approach for this type of detection in previous researches is based on the principal score (PS) of each sliding window by computing top eigenvalues of the correlation matrix, e.g. the Lanczos algorithm. This paper demonstrates the phenomenon of principal score degeneration for large data set, and then mathematically and practically prove all current PS-based methods are bound to fail for anomalous correlation detection on large-scale streaming data, even when the number of correlated anomalies grows with the data size at a reasonable rate; in reality, anomalies tend to be the minority of the data, and this issue can be more serious. We propose a framework with two novel randomized algorithms rPS and gPS for better detection of correlated anomalies from large streaming data of various correlation strength. The experiment shows high and balanced recall and estimated accuracy of our framework for anomaly detection from a large server log data set and a U.S. stock daily price data set in comparison to direct principal score evaluation and some other recent group anomaly detection algorithms. Moreover, our techniques significantly improve the computation efficiency and scalability for principal score calculation.

## I. INTRODUCTION

Anomaly detection is an active research topic of long history, which aims to discover abnormal pattern that deviates from the vast common. Although a majority of algorithms find anomalies as individuals [1], some other algorithms aim at discovering *anomaly groups* rather than single individuals, for example, [2], [3], [4], [5] and [6]. The individuals in each anomaly group need not appear to be anomalous by themselves, but collectively they become noticeably different from others. These models are classic, either define an anomaly score to search for a subset with large score, or devise a probabilistic model that finds observations with small probabilities. In this paper, we refer to an anomaly group as an *anomaly set*.

Our research is about one type of group anomaly detection, called *correlated anomaly detection* (CAD), and we focus on data streams. CAD *assumes* the normal data entries in data streams are weakly correlated or at least not strongly correlated for most of the time, and strong correlations can be considered unlikely and anomalous. A group of data entries are *correlated anomalies* if they have strong internal correlations. We only consider Pearson's correlation.

Correlated anomalies are relatively less studied in literature than classic group anomalies, they actually have **interesting industrial applications**. For example, the botnet detection is to find correlated intrusion detection [7], or malicious server visits from distributed denial of service (DDOS) attackers, price scraping system, and ill-will web crawlers [8] [9]. For another example, the correlated price changes in stock markets is an indicator of potential financial events [10]. Furthermore, anomalous correlation is applied in studying chemical process [11], industrial process [12] and sensor data [13]. We discuss two problems in detail for intuitive understanding of what CAD problems we are dealing with,

- *Correlated server requests*. Suppose a web server receives requests from $N$ different IPs for a time window (e.g. 1 hour) where $N$ is a relatively large number (e.g. tens of thousands or even higher). Suppose these $N$ IPs have requested $M$ different files on the server during this period, then we can construct a $M \times N$ matrix $\mathbf{R}$ where $\mathbf{R}(i,j)$ is the number of times the $j$th IP request the $i$th file within this hour. Let $\mathbf{R}(\cdot, j)$ denote the $j$th column of $\mathbf{R}$. If we randomly sample a few columns from $\mathbf{R}$, they are usually weakly correlated or at least strongly correlated, because normally we can assume requests from different IPs are independently made by different people following different visit patterns. Our objective is to determine if there is anomalously strong correlation among some IPs, i.e. if there exist a subset of IPs indexed by $j_1, \ldots, j_k$ s.t. the column vectors $\mathbf{R}(\cdot, j_1), \ldots, \mathbf{R}(\cdot, j_k)$ have strong internal correlation. It is possible these IPs are used by some coordinated distributed automatic programs (e.g. botnet) for malicious purposes like information scraping or DDOS attacks. Figure 1(a) demos an example of a confirmed botnet actually captured by our algorithms.

- *Correlated stock fluctuations*. Suppose we have $N$ stocks in the market. For each stock, its price changes within a time window (e.g. one month) can be converted into a vector of price-change percentages. For example, if the prices are $19.2, $19.7, $18.5, $18.3, $19.0, ..., then the price-change percentages are 2.6%, -6.2%, -1.1%, +3.8%, ... Suppose every stock have $M + 1$ price records in this time period, then we can have a $M \times N$ matrix $\mathbf{F}$ such that $\mathbf{F}(i,j)$ is the $i$th price-change percentage for the $j$th stock. Similar to above, our objective is to determine if there is a subset of stocks whose price changes are correlated. Such strong correlation is actually rare given a sufficiently long period and sufficiently many price changes, even for stocks from the same industry. For example, we tested correlations for price changes of stocks in gold industry for 12000 30-day time windows; we found on average two gold stocks show strong price-change correlation (above threshold 0.7) for about 45 of these time windows, illustrated in Figure 1(b). Stocks from other industries have similar results. Therefore, these strong correlations can be a good indicator of a financial event.

* Equal contribution to this research.

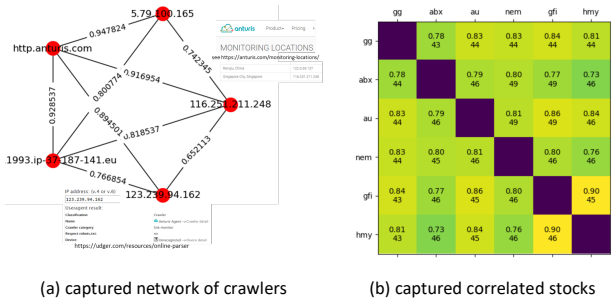

(a) captured network of crawlers   (b) captured correlated stocks

Figure 1. **(a)** A crawler network identified from a large server log dataset. The network comes from a company named Anturis, where some IPs are listed on their website (so we confirm these IPs come form this company). This botnet causes multiple alters when our algorithms run through the server log data stream. The edge weights are average request correlations between two IPs in these alerts. **(b)** The labels along both sides of the matrix are stock symbols whose full name can be found online. In each matrix cell, the integer at the bottom is the number of times the price changes of the two stocks have strong correlation (higher than 0.7) among 12000+ 30-day time windows, the decimal at the top is the average of these price-change correlation if they are above the threshold. This illustrates that strong price-change correlation is rare and can be considered anomalous.

*CAD is technically different* from classic detection because data entries are not considered anomalous if they do not exhibit strong internal correlation, even if they are "unlikely" to be observed by a classic detector. Therefore, classic models mentioned in the first paragraph are generally not employed in current CAD researches; rather, current CAD researches are mainly based on *Principal Component Analysis* (PCA), and almost all researches on CAD deal with streaming data and monitor the anomalies on the fly; however, as far as we know, recent algorithmic development is insufficient to better catch up with the ever-growing data scale. Recent algorithms for detecting anomalies from streaming data, such as [14], [15], [16], [17], [18], are only good for extraction of individual anomalous data vectors from the streams, and also they are not targeting correlated anomalies.

We now overview how *"big data" brings modern challenges to correlated anomaly detection*. A canonical approach is to set a sliding window on the data stream, converting data entries in the window into a feature matrix, and run PCA on the feature matrix. If the *principal score* (the eigenvalue associated with the principal component divided by the number of entries) is higher than a threshold, say 0.7, it means plenty of data points are scattered around the rotated axis defined by the principal component and thus correlated. We find the challenges to extend this approach to large-scale streaming data is two-fold.

- *First challenge*, the anomalous correlation is significantly weakened. On one hand, a time window of a large data stream could contain a large amount of uncorrelated normal data entries that overwhelm the anomalous correlation regardless how strong it is. Later Figure 2 illustrates even under realistic assumption like the anomaly data take up a fixed portion of the total data, the principal score still asymptotically degenerates to 0.5 when the data size grows. A rigorous discussion and a proof is provided in Section III.B. On the other hand, anomalies are trying to hide themselves, which makes the situation even worse. For example, a botnet can simulate human-like visits on a server where each individual bot looks very normal, hard to be discovered by single-anomaly detectors, but altogether they still impose a heavy burden [19]. We refer to this problem as *principal score degeneration*, and we propose this issue needs to be prevented before a PCA-based approach can actually be extended to large-scale streams.

- *Second challenge*, we need to reduce the algorithmic time complexity. Principal score computation is intense with its sub-cubic time complexity and might not a feasible solution to monitor large-scale streaming data. What is worse, industrial application could require simultaneous real-time analysis with different window sizes from 15 mins to half day. The current PCA-based methods either need to skip a large part of the streaming data [8], which is not applicable for online analysis, or focus on speeding up the eigenvalue algorithms by a small order [19, 20].

Consequently, we believe current researches are insufficient for the purpose of mining correlated anomalies from large data streaming. This paper proposes *two related algorithms* to meet the challenge. *The first algorithm* comes with the idea that we could sample from the data with a distribution that favors potential correlated anomalies based on "strong anomaly strength" assumption to be introduced later. The sampling not just makes the algorithm more sensitive to "hidden" correlations, and meanwhile it reduces unnecessary computation. We name both the algorithm and the principal score it generates as rPS (Randomized PS). *The second algorithm* models the correlation matrix from a novel generative perspective. It is completely different from the typical mix of LDA and Gaussian mixture [21] for classic group anomaly detection. To our best knowledge, we could be the first to attempt a generative model for the correlated anomaly detection. The second algorithm does not make strong assumptions and is capable of mining correlated anomalies even when the first algorithm cannot properly detect them. We name the algorithm and the principal score it generates as gPS (Generative PS). The two algorithms rPS and gPS *make a good partner* where the output of the first algorithm can be used to initialize the second algorithm.

We perform experiments on two datasets. The first is a large server log data set of 315 million log entries; the second is a stock price dataset of 30+ years daily price data. The experiments show the proposed algorithms can achieve much better performance in terms of recall and estimated accuracy, in comparison to PCA-based methods [12, 19] and two additional models [3, 6] as representatives of afore-mentioned classic group anomaly detection methods (to show they do not work for CAD even with tweaks as best as we can), and the computation time is dramatically reduced.

We summarize the *main contribution* of this paper as the theoretical discussion and proof of *principal score degeneration*, and the two algorithms to overcome this issue, as well as the experiments on two real-world datasets. In Section III, we start with theoretical discussion that points the way for development of two algorithms rPS and gPS. In Section IV, we first observe qualitative results consistent with the discussion of Section III, and then conduct quantitative experiments to compare performance.

## II. RELATED WORKS

Group anomaly detection is more difficult than individual anomaly detection because it involves exploration of underlying data structure and the design of unlikeliness measure for observations. For example, [2] trains "boxes" in the feature space for each data stream from multiple "normal" streams and then merge nearby boxes. The distance from a data point to the closest box is its anomaly score. [6] detects anomaly groups on a graph. It defines an anomalous "metric" for each node and edge. The algorithm starts with all nodes of the graph as an anomaly group, and for each iteration the algorithm tries to remove one node from the group that best increase the group's anomaly score. The method is simple and intuitive, and its simplicity leads to a mathematical error bound. In comparison, [3] builds a generative model that combines Latent Dirichlet Allocation (LDA) and Gaussian mixture, and an observation with small probability under the model is naturally considered an anomaly. Other researches like [4, 5, 22] etc. generally follow the two schemes.

Among all above methods we reviewed, we find only [6] could be applied to correlated anomaly detection, even correlation is not considered by design. The algorithm starts with the entire graph, then in each step it keeps removing the node with the least sum of edge weights plus node weights. The sum of edge weights and node weights of the remaining subgraph divided by the number of nodes is the anomaly score. In this way, it effectively searches for a denser subgraph of high anomaly score, thus noticeable correlation could be expected if it can truly find a dense subgraph.

Correlated anomaly detection (CAD) falls in the category of group anomaly detection, but its approaches in previous researches are very different from those for the classic group anomaly detection, since in this case the unlikeliness is solely measured by correlation. The primary technique for CAD is PCA [8, 19, 20, 23, 24], technically computing the top eigenvalues and their eigenvectors. The principal score is calculated by dividing the largest eigenvalue by the data size; it provides a unified correlation score supported by well-developed statistical and numerical theory. PCA can also be applied in classic anomaly detection for a different purpose -- dimension reduction [25, 26].  Although CAD typically involves streaming data since the beginning, none of the above-mentioned works can directly apply to large-scale streams, due to the principal score degeneration. Meanwhile, as far as we know, the best time complexity reported from previous works is sub-cubic [12, 19, 27], still intensive for large-scale monitor and involving much unnecessary calculation that could be avoided.

## III. MODEL AND INFERENCE

### A. Problem Statement

Given a *data matrix* $\mathbf{X} = (\mathbf{x}_1, \dots, \mathbf{x}_n)$ where each column $\mathbf{x}_i$ is a data vector and $n$ is very large, we intend to discover a subset of the data vectors in $\mathbf{X}$, called *anomalies*, with strong internal anomalous correlation, that is, most pairs of the detected anomalies have strong pairwise positive or negative correlations. Define *correlation matrix* $\mathbf{P}$ as an $n \times n$ matrix consisting of pairwise correlations between every pair of data vectors in $\mathbf{X}$. The diagonal elements of $\mathbf{P}$ are clearly all 1 because self-correlation is always 1. We make all correlations in $\mathbf{P}$ lie in [0,1] by taking absolute values in order for $\mathbf{P}$ to be a positive semi-definite matrix as well as a non-negative matrix. Note this operation of taking absolute values will not affect the generality of our detection theory and algorithm. After detection of anomalies, the sign of pair-wise correlations can be recovered by re-calculate the pairwise correlations among the anomalies; since the number of detected correlated anomalies is typically small, this recovery will be of negligible cost. In other cases, if it is intended to only discover positive correlated anomalies, we can simply do another operation of setting negative values in $\mathbf{P}$ as zero; if it is intended to only discover negative correlated anomalies, we can set positive values in $\mathbf{P}$ as zero and then analyze $-\mathbf{P}$ instead. Therefore, in a word, no matter which case, we can consider $\mathbf{P}$ as a non-negative matrix without loss of generality of our theory and algorithms.

Denote the largest eigenvalue of $\mathbf{P}$ as $\lambda_1$, then $\rho(\mathbf{X}) = \frac{\lambda_1}{n} \in$ [0,1] is called the *principal score*. The anomaly detection task is based on following assumption.

**Assumption 1**. Normal data entries have weak or at least non-strong correlation. Therefore, when $\rho(\mathbf{X})$ is closer to 1, then it implies more likelihood that $\mathbf{X}$ may contain correlated anomalies.

The objective is thus to compute $\rho(\mathbf{X})$. If $\rho > \tilde{\rho}$ for some threshold $\tilde{\rho}$ (say $\tilde{\rho} = 0.7$ ), we should ***emphasize*** lowering threshold $\tilde{\rho}$ is dangerous and generally not a good practice because of the "weak correlation" in Assumption 1, and $\tilde{\rho} = 0.7$ is an intuitively reasonable value, and it is also supported by our experiment results (see Table 1, lowering threshold causes many false positives). If $\rho > \tilde{\rho}$, we then consider anomalous correlation in the data is possible, and an alert can be raised for human attention. Along with calculation of $\lambda_1$, the principal component (eigenvectors of $\lambda_1$) can be found, and any data entry with strong correlation to it are considered as correlated anomalies. For streaming data, a usual practice is to either divide the data stream into batches or use a sliding window.

### B. Principal Score Degeneration

We now show the intrinsic problem of using principal score $\rho$ for detecting correlated anomalies from large-scale streaming data. A time window may contain tens of thousands of data entries where usually only a portion of them are anomalous and show unusual strong correlation. They will be overwhelmed by other "large number of" normal data entries when computing $\rho$. The magnitude of the anomalies does not help the detection because correlation is insensitive to magnitude, e.g. the correlation between two vectors $\mathbf{x}_1 = (1,0,2,2,0,1)$ and $\mathbf{x}_2 = (0,0,1,2,1,1)$ is the same as the correlation between $\mathbf{y}_1 = (10,0,20,20,0,10)$ and $\mathbf{y}_2 = (0,0,10,20,10,10)$. This is unlike classic single-anomaly detection where large vectors can be more easily detected. For a concrete example, in case of detecting malicious server visits (bot detection), even the attackers try to impose heavy workload to bring the server down, because they are the "small number" in comparison to the total number of users, $\rho$ would still be nowhere near the alert threshold. It is also imaginable the problem is more serious if the data size grows. We call this phenomenon as *principal score degeneration*. Although this is intuitively true, a rigorous proof is difficult and non-trivial.

In practice, it is of course reasonable to assume the anomaly set could grow with the data size. For example, if the server is capable of processing more requests, an attacker could invest more computers. It is interesting to ask 1) if the anomaly sets could still be overwhelmed if they grow with the data size; 2) which value $\rho$ is expected to degenerate to?

One contribution of this paper is we found a rigorous mathematical proof to answer above questions under the following Assumption 2 which real-world datasets generally satisfy. We *discover* theoretically $\rho$ will be overwhelmed even if the anomaly set grows almost at the same order as the data size, and the actually calculated $\rho$ (not just its expectation $\mathbb{E}\rho$) will be guaranteed to degenerate to the background correlation $\mu$ (the average of all correlations excluding those anomaly-anomaly correlations). The discovery strongly implies it is inappropriate to directly apply any principal-score based algorithm on a large data set. In the proof, we assume the whole data set is of size $n + k + 1$ where $k$ is the number of correlated anomalies in the data set. In the proof and future discussion, abbreviation "s.t." stands for "such that", and "w.r.t." stands for "with respect to".

**Assumption 2**. The average of the correlations between normal data entries in **X** do not exceed the alert level $\tilde{\rho}$. This is a very loose assumption which real-world datasets will generally satisfy.

**Proposition 1**. *Principal Score Asymptotic Upper Bound*. Given data matrix **X** of $n + k + 1$ columns with $k = O(n^m), m \in [0,1)$ anomalies of high mutual correlation, and assume all other absolute correlations are independent random numbers in [0,1] with mean $\mu$ and finite variance $\sigma^2$. If we fix $k$ and let $n \to \infty$, then $\rho(\mathbf{X}) \leq \mu + o\left(\frac{1}{n^m}\right)$ for any $0 \leq m < \min\left\{\frac{1}{2}, 1 - m\right\}$ with probability arbitrarily near 1, and $\mathbb{E}\rho \to \mu$. Here $O$ and $o$ are standard big-O and little-o notation.

*Proof*. Let **P** be the correlation matrix, and $s_i$ be the sum of the $i$th row of **P**. Since **P** is non-negative, then Perron-Frobenius theorem implies $\frac{\min s_i}{n} \leq \rho \leq \frac{\max s_i}{n}$. By assumption, $s_i \leq 1 + k + nX$ where $X$ is a random variable of the same distribution from which the random correlation is drawn, then we have

$$\frac{\mathbb{E}[s_i]}{n} = \frac{1 + k + n\mathbb{E}X}{n} \to \mu, i = 1, \ldots, n \Rightarrow \mathbb{E}\rho \to \mu$$

Let $\tilde{s}_i$ be the sum of the $i$th row excluding the diagonal and the correlations involving the $k$ anomalies, then $s_i \leq k + 1 + \tilde{s}_i$. If $\mathbf{x}_i$ is non-anomaly, then $\tilde{s}_i$ is the sum of $n$ i.i.d. random variables; otherwise $\tilde{s}_i$ is the sum of $n + 1$ i.i.d. random variables. We discuss for non-anomaly without loss of generality.

Given any $\epsilon > 0$, let $x$ be a number determined by equation $\epsilon = \frac{1}{\sqrt{2\pi}} \frac{n}{x} e^{-\frac{x^2}{2}}$. Let $n_{r,S}$ be the number of $r$-tuples of row sums at least $S = n\mu + x\sqrt{n\sigma^2}$. As $n \to \infty$, we have

$$\mathbb{E}[n_{r,S}] = \binom{n}{r}\left(\mathbb{P}\left(s_i \geq n\mu + x\sqrt{n\sigma^2}\right)\right)^r$$

$$\sim \binom{n}{r}\left(\int_x^{+\infty} \frac{1}{\sqrt{2\pi}} e^{-\frac{x^2}{2}}\right)^r$$

$$\sim \frac{n!}{r!(n-r)!n^r}\left(\frac{n}{\sqrt{2\pi}x} e^{-\frac{x^2}{2}}\right)^r \to \frac{\epsilon^r}{r!}$$

where $\frac{s_i - n\mu}{\sqrt{\sigma^2 n}}$ is standard normal by central limit theorem and $\int_x^{+\infty} \frac{1}{\sqrt{2\pi}} e^{-\frac{x^2}{2}} \sim \frac{1}{\sqrt{2\pi}x} e^{-\frac{x^2}{2}}$. Let $n_S$ be the number of row sums at least $S$, then $n_{r,S} = \binom{n_S}{r} \Rightarrow \mathbb{E}[n_S(n_S - 1)\ldots(n_S - r + 1)] = \epsilon^r$. We can check the coefficient of $n_S^i, i = 1, \ldots, r$ in the product inside the expectation equals $\begin{Bmatrix} r \\ i \end{Bmatrix}$ (2nd kind Sterling number). This implies the $r$th moment $\mathbb{E}[n_S^r]$ of $n_S$ is exactly the $r$th moment of a Poisson random variable for any $r = 1, \ldots, n$. By Carleman's condition[28], $n_S$ obeys Poisson distribution parameterized by $\epsilon$, implying

$$\mathbb{P}\left(\frac{\max \tilde{s}_i}{n} < \mu + \frac{x\sigma}{\sqrt{n}}\right) = \mathbb{P}(n_S = 1) = e^{-\epsilon}$$

Finally, we can in particular let $\epsilon = \frac{1}{\sqrt{4\pi \log n}}$, then it is easily seen

$$\mathbb{P}\left(\frac{\max s_i}{n} < \mu + \sigma\sqrt{\frac{2\log n}{n}} + \frac{k+1}{n}\right) \approx 1 \text{ when } n \text{ is large. With } \rho \leq \frac{\max s_i}{n}, \frac{k+1}{n} = O\left(\frac{1}{n^{1-m}}\right) \text{ and } \sqrt{\frac{\log n}{n}} = o\left(\frac{k+1}{n}\right) \text{ when } m \in \left(\frac{1}{2}, 1\right), \text{ we have } \rho \leq \mu + o\left(\frac{1}{n^m}\right) \text{ for any } 0 \leq m < \min\left\{\frac{1}{2}, 1 - m\right\} \text{ with probability arbitrarily near 1. } \blacksquare$$

Proposition 1 affirms the eventual failure of methods based on just calculating or approximate the principal score of the entire correlation matrix, as those mentioned in related works: it literally means even if the number of correlated anomalies has a reasonable growth rate with strong anomalous correlation, they will be "ignored" when provided a large amount of other normal data, and the principal score has no chance to exceed the mean of background correlation plus a small perturbation. We note Proposition 1 only gives the upper bound, so the actual case could be worse. A numerical illustration is given in Figure 2. Originally about 50% of the data are anomalies with average correlation 0.85. We randomly add both anomalous and normal data. The first chart keeps $k = n^{0.8}$, which falls in the domain of Proposition 1. The second chart keeps $k = 0.2n$, growing at the same order as $n$, yet the principal score degenerates anyway.

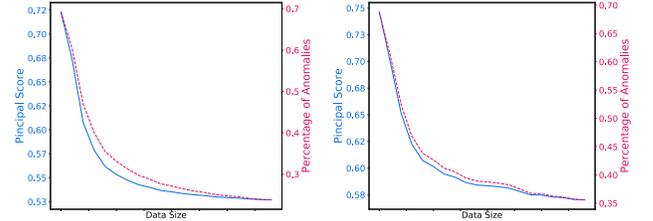

Figure 2 Degeneration of Principal Score. The growing rate of the left is $k = n^{0.8}$ (falls in the range specified in Proposition 1), of the right is $k = 0.2n$.

## C. Randomized PS

Our first randomized algorithm tries to resolve above problem and extend the principal-score based method to large streaming data that satisfies "strong anomaly strength" assumption defined below. Such anomalies are of big concern in applications like bot detection or intrusion detection. Our technique falls in the category of randomized algorithms [29-31] of the numerical computation society. Although there have been many such algorithms for PCA or SVD and techniques like interpolation, QR factorization, subspace iteration, etc., their goal is to preserve the principal score and principal components, so the approximated eigenvalues of the correlation matrix can be as accurate as possible. However, extreme accuracy is not very needed in our application, and in contrast our randomized algorithm does the opposite: it needs to noticeably increase the principal score. We now introduce a second proposition, which reveals what conditions can prevent the principal score from being overwhelmed as seen in Proposition 1, and also serves as the *correctness* for later proposed sampling technique. The proof is based on the random matrix theory [32].

**Proposition 2**. *Asymptotic Concentration of Correlation with Anomalies*. Suppose $k^2 - k$ anomalous correlations in an $n \times n$ correlation matrix **P** are random numbers in [0,1] with $\tilde{\mu}$ as its mean, and all other correlation are independent random numbers in [0,1] with mean $\mu$ s.t. $\mu < \tilde{\mu}$. As $n \to \infty$, if $\frac{k}{n} \to \varphi$ for some constant $\varphi \in (0,1)$, then $\rho$ concentrates around $\mu + (\tilde{\mu} - \mu)\varphi^2 \pm O\left(\frac{1}{\sqrt{n}}\right)$.

*Proof*. Use random matrix theory. With large $n$, every element above the diagonal of **P** can be viewed i.i.d. generated by two steps: with probability $\varphi^2$ it is drawn from a distribution with mean $\tilde{\mu}$, and with probability $1 - \varphi^2$, it is drawn from a distribution with mean $\mu$. Thus, the mixed distribution has mean $\varphi^2 \tilde{\mu} + (1 - \varphi^2)\mu$, and $\frac{\mathbf{P}}{\sqrt{n}}$ is a Hermitian Wigner matrix (diagonal is constant in our case, elements above the diagonal are i.i.d.). By its property [32], the top eigenvalue of **P** concentrates around $(\varphi^2 \tilde{\mu} + (1 - \varphi^2)\mu)n \pm O(\sqrt{n})$, so the principal score $\rho$ concentrates around $\mu + (\tilde{\mu} - \mu)\varphi^2 \pm O\left(\frac{1}{\sqrt{n}}\right)$. ∎

In plain words, $\varphi = \frac{k}{n}$ is the percentage of the anomalies, $\tilde{\mu}$ is the mean correlation between anomalies, and $\tilde{\mu} - \mu$ is the difference between the average anomalous correlation and the average normal correlation. Proposition 2 implies *two conditions* for the principal-score based method to work well for large data set: 1) the number of anomalies keeps taking a sufficiently large percentage, 2) and the anomalous correlation is sufficiently distinguishable from the normal correlation. Satisfy both conditions, then $\rho(\mathbf{X})$ will be theoretically above $\mu$ by a noticeable non-diminishing quantity. It is also not hard to verify the two conditions are not only sufficient, but also necessary, because we have proved in Proposition 1 that any sub-$n$ growth order of anomalies will eventually make $\rho(\mathbf{X})$ no bigger than $\mu$.

The second condition is an intrinsic property of the data set, and we cannot change it; but we could make the data satisfy the first condition through sampling, if the anomalies can have higher probability to be sampled. We should in addition note although Proposition 2 suggests an apoptotic constant $\frac{k}{n}$ is theoretically enough, given the result of Figure 1, we should allow sampling more anomalies.

**Definition 1**. Let $A = \{\mathbf{x}_{i_1}, \ldots, \mathbf{x}_{i_k}\}$ be a set of correlated anomalies in **X**, define $\phi(A) = \frac{\sum_{j=1}^{k} \|\mathbf{x}_{i_j}\|_p}{\sum_{i=1}^{n} \|\mathbf{x}_i\|_p}$ for some $p \geq 1$ as the *anomaly strength* of $A$ in **X** (i.e. with respect to the $p$-norm). Note anomaly strength defined in this way is independent from the size of an anomaly set; a small anomaly set (small $k$) might have strong strength if they have large-magnitude data vectors; a large anomaly set can have strong anomaly strength even if each anomaly consists of normal quantities.

**Assumption 3**. *Strong anomaly strength assumption*. Big quantity in a data vector tends to be anomalous, and an anomaly set has sufficiently strong strength.

Above assumption is not unusual in practice and is a primary concern for many cases. In bot/botnet detection, anomaly strength is equivalent to traffic volume, and above assumption covers the case when abnormal visits impose a noticeable workload on the server, either to slow down the server, or to intensively collect information. In stock/exchange market, above assumption covers the case when many stocks/currencies show small correlated fluctuations, or a few stocks/currencies show huge correlated fluctuations; both have meaningful implication for financial market [33].

We propose random sampling with replacement according to $p$-norm for $p \geq 1$ from the data entries to form a new matrix on which the principal score is computed. Each data entry **x** has probability $\frac{\|\mathbf{x}\|_p}{\sum_{i=1}^{n} \|\mathbf{x}_i\|_p}$ to be sampled. We refer to both this technique and the principal score evaluated from the sample as rPS (Randomized PS). Under above-mentioned assumption, if the anomaly set has strength $\phi$, then this ensures the number of anomalies are expected to make up ratio $\phi$ of the sampled data, regardless how large the anomaly set is. By Proposition 2, the principal score is likely to be around $\mu + (\tilde{\mu} - \mu)\phi$ and will be noticeably above the normal data mean correlation if $\phi$ is sufficiently large.

The parameter $p$ controls sampling sensitivity to large quantities and how strong the strength should be, as stated in Assumption 3, such large quantities tend to be anomalous. However, larger $p$ might increase the chance of false alarm by potential sampling multiple times the same large vector. The other parameter is sampling ratio (sampled size divided by the data set size), denoted by $r$. The parameter tuning is shown in section IV.C. We can apply the Lanczos algorithm with sub-cubic time complexity to evaluate the principal score. The time complexity of rPS is at the same order of underlying PS algorithm applied on the sample, but with a much smaller coefficient $r^2 \ll 1$.

*D. Generative PS*

We have two main *motivations* to bring up a second algorithm. *First*, sometimes it is possible the anomaly strength in data is weak, for example, crawlers could collect information from a popular website at a mild visiting rate in order to protect themselves from exposure. For the stock price dataset, unlike the server logs data set where anomalies prevail, it is not usual for a large number of stocks to fluctuate in a correlated way. When "strong anomaly strength" assumption as in Assumption 3 does not hold, it is not easy to directly find useful information for sampling. *Secondly*, we find from our experiments that rPS has a noticeable chance to raise false alerts and makes incorrect identifications.

The design of gPS is based on the same idea as rPS to favor principal score evaluation on potential anomalies. In the proof of Proposition 2 we can see the correlation matrix can be viewed as being "generated". Inspired by this, we propose a generative view of correlation matrix to infer the anomaly sets.

**Assumption 4**. Each anomaly set has strong internal correlation among themselves but have much weaker external correlation to other data entries. We believe this is generally true for all data sets. It might not be hard for anomalies to have "camouflage" by simulating the quantities of normal data entries, but it would be very difficult for them to well simulate realistic correlations.

We refer to the following model and its principal score as *gPS* (Generative PS). We assume there are $\mathfrak{k}$ anomaly sets in the data, where each anomaly set is associated with two parameters $a_i, b_i, i = 1, \dots, \mathfrak{k}$ s.t. $a_i + b_i \geq 1$ and $\frac{a_i}{a_i+b_i} \geq a_i$ for some $a_i \geq a$ for some $a > 0.5$. The first restriction keeps the beta distribution to have a valid shape for our purpose; the second constraint $a$ is the average correlation between two data entries in an anomaly set, which intuitively should be a value near the threshold $\tilde{\rho}$ and based on Table 2 we choose $a = 0.75$. There is one additional pair of parameters $a_{\mathfrak{k}+1}, b_{\mathfrak{k}+1}$ for non-anomalies such that $a_{\mathfrak{k}+1} + b_{\mathfrak{k}+1} \geq 1$ and $\frac{a_{\mathfrak{k}+1}}{a_{\mathfrak{k}+1}+b_{\mathfrak{k}+1}} \leq 0.5$. $\mathfrak{k}$ can be treated as a small positive integer constant independent of data size since intuitively there should not be many anomaly sets in most situations.

Now define a vector $\mathbf{z} = (z_1, \dots, z_n)$ where $1 \leq z_l \leq \mathfrak{k}+1$ indicating if data $\mathbf{x}_l$ is in one of the $\mathfrak{k}$ anomaly sets, and an element $\mathbf{P}(i,j)$ in the correlation matrix above the diagonal is viewed as been drawn from beta distribution parametrized by $a_i, b_i$ if its corresponding data $\mathbf{x}_i, \mathbf{x}_j$ are both in the $i$th correlated subset for some $1 \leq i \leq \mathfrak{k}$ (anomalies); otherwise it is viewed as been drawn from a beta distribution parametrized by $a_{\mathfrak{k}+1}, b_{\mathfrak{k}+1}$ (non-anomalies).

Beta distribution is a widely used continuous distribution over [0,1] with flexible shape, and the restriction of the parameters ensure its unimodality. For simplicity, let $m_i$ be the number of correlations for the $i$th group, and $m_{\mathfrak{k}+1}$ be the number of all other correlations, and $\omega_{i,l}, l = 1, \dots, m_i, i = 1, \dots, \mathfrak{k}+1$ be the corresponding correlations. Therefore, the likelihood of the correlation matrix is

$$L(\mathbf{P}) = \prod_{i=1}^{\mathfrak{k}+1} \left( \left( \frac{\Gamma(a_i+b_i)}{\Gamma(a_i)\Gamma(b_i)} \right)^{m_i} \prod_{l=1}^{m_i} \omega_{i,l}^{a_i-1}(1-\omega_{i,l})^{b_i-1} \right)$$

$$\Rightarrow \ln L(\mathbf{P}) \propto \sum_{i=1}^{\mathfrak{k}+1} \left( (a_i-1) \sum_{l=1}^{m_i} \ln \omega_{i,l} + (b_i-1) \sum_{l=1}^{m_i} \ln(1-\omega_{i,l}) + m_i(\ln \Gamma(a_i+b_i) - \ln \Gamma(a_i) - \ln \Gamma(b_i)) \right)$$

Solve for the derivative of each $a_i$ and $b_i$, we have the following iteration formulas where $\psi$ denotes digamma function, $a_i^{(s)}, b_i^{(s)}$ denote the parameter values at the end of the $s$th iteration. The afore-mentioned parameter constraints are enforced during the following updates

$$a_i^{(s+1)} = \psi^{-1}\left( \psi(a_i^{(s)}+b_i^{(s)}) + \frac{\sum_{l=1}^{m_i} \ln \omega_{i,l}}{m_i} \right)$$

$$b_i^{(s+1)} = \psi^{-1}\left( \psi(a_i^{(s)}+b_i^{(s)}) + \frac{\sum_{l=1}^{m_i} \ln(1-\omega_{i,l})}{m_i} \right)$$

After updating beta parameters, the update of $z_i$ is specified by a maximization over a finite support,

$$z_i = \arg \max_{z_i=1,\dots,\mathfrak{k}+1} \ln L(\mathbf{P})$$

We use the anomalies detected by rPS as the initialization of gPS for $\mathbf{z}$; then the beta parameters are initialized in a way such that the distribution mean is the average correlation of each anomaly set while the afore-mentioned constraints are enforced. After convergence, the $\mathfrak{k}$ potential anomaly sets are labelled in $\mathbf{z}$ and we can compute the principal score for each of them. One interesting and distinct advantage of gPS is its capability to cluster anomalies with strong internal correlations, so the detected anomaly set is highly trustable. An example in Figure 3, where top left brighter square is clustered by gPS, surrounded by a region where rPS also detects other anomalies with weaker correlations and potential errors. The possible reason for the lesser accuracy of rPS is that principal score identifies an anomaly if the correlation between a data entry $\mathbf{x}_i$ and the principal component is above a threshold; however, this correlation mathematically equals $\mathbf{q}_i(1)\sqrt{\rho}$ where $\sqrt{\rho}$ is not a small quantity, and $\mathbf{q}_i(1)$ is the first value of the $i$th principal component with no known guarantee to be a small quantity. We define the anomalies detected by both rPS and gPS as the *core anomalies*, and other detected anomalies as *suspicious anomalies*. A suspicious anomaly can go through further analysis that uses historical data to better confirm its identity.

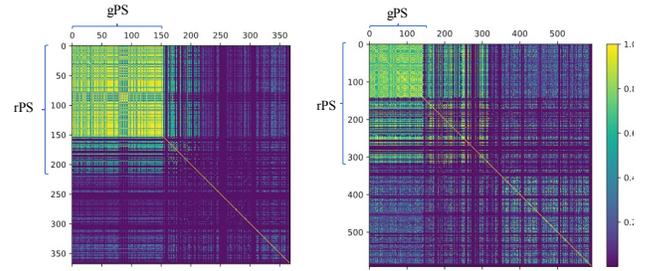

(a) Visualization of two correlation matrices from real data showing gPS detects "core" anomalies. Rows and columns are permuted for better visualization where anomalies are placed at the top-left corner.

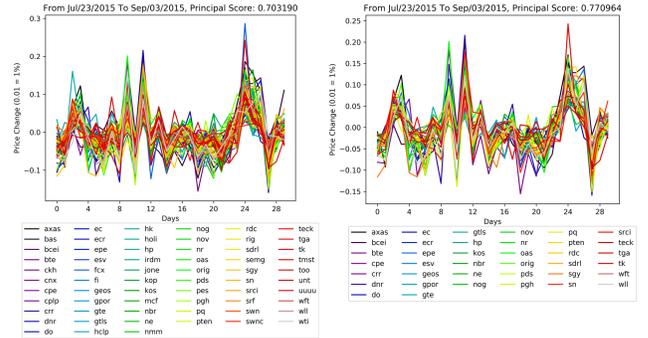

(b) Anomalous stock price fluctuations identified by rPS (left) and gPS (right) for the same 30-business-day time window. gPS recognised a nearly "subset" of what rPS detects, with a higher correlation. The legend labels are stock symbols whose full stock name can be found online.

Figure 3 gPS makes more conservative decision than rPS.

At last, we prove the quadratic time complexity of gPS in Proposition 3.

**Proposition 3**. *gPS Time Complexity*. Each iteration of the updates for the generative model gPS has at worst $O(\varphi n^2)$ time complexity, where $\varphi$ is the percentage of anomalies.

*Proof*. The overwhelming time usage is for updating $\mathbf{z}$. We note updating $z_i$ requires computing the sum " $a_i \sum_{i=1}^{m_i} \ln \omega_l + b_i \sum_{i=1}^{m_i} \ln(1-\omega_l)$" for each $1 \leq i \leq \mathfrak{k}$ in the likelihood for those addends s.t. 1) correlation $\omega_l$ is between $\mathbf{x}_i$ and another data entry; 2) $i \in \{z_i^{(t)}, z_i^{(t+1)}\}, 1 \leq i \leq \mathfrak{k}$, i.e. the re-computation is only necessary for at most two correlated anomaly sets. In this case, we at most need to compute the sum of $\varphi n$ "$\omega_l$" and multiply it to the corresponding beta parameters. Since $k$ is a small fixed number, each round of updating entire **z** needs $O(\varphi n^2)$ complexity. ∎

*E. Put Together*

In previous discussion we described two randomized models rPS and gPS for correlated anomaly detection and defined those detected by rPS as suspicious anomalies and those detected by both rPS and gPS as core anomalies.

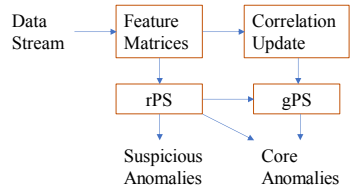

Figure 4 The Process of Our Framework of Correlated Anomaly Detection

Now Putting everything together, a proposed data processing flow for correlated anomaly detection is shown in Figure 4. Two algorithms rPS and gPS are combined into a framework and they make a partner to achieve the goal of efficient online detection from large streaming data. A sliding window moves through the data stream and converts data within the time window to feature matrices. Both algorithms require computing correlation matrix. The naïve computation costs cubic time complexity; fortunately, for online streams we can use the recently developed online formula in [19] with quadratic time complexity, so that we do not recompute a majority of the correlations after the time window slides.

## IV. EXPERIMENTS AND EVALUATION

We experiment on two large datasets with time attributes, so they can simulate streaming data. ***The first dataset*** is a 4-month ecommerce web server log dataset, which contains 315,688,764 Apache access log entries and visits from 2,519,022 distinct IPs. The server is known to be attacked twice and constantly harassed by web crawlers possibly collecting real-time price information. Given a time window, e.g.1 hour, the log data can be converted to a URL-IP matrix where each column represents the number of times an IP visits an URL. This is a very noisy dataset where the average background correlation is higher than 0.5. ***The second dataset*** contains all daily stock price history of about 7200 U.S. stocks [34], and we aim to capture correlated price changes.

We compare our algorithms to algorithms based on direct principal score evaluation [19, 20], referred to as "direct PS". These algorithms have the same nature as the principal score evolution from classic PCA, although the recent methods do make improvement to speed up evaluation, they do not recognize the issue of principal score degeneration. Specifically, we use the implementation in [19].

We also compare to Fraudar, a recent group anomaly detection model by [6]. Among all researches we reviewed in Section II, this is the only unsupervised model that could be applied to our problem. We use their recommended setup, except for we do not down-weight nodes with large degrees. For social network or spam review detection, large-degree nodes can indicate popularity rather anomaly, but our cases are obviously different. For the first dataset, we experiment it on the URL-IP bipartite graph. On the second dataset, we are only able to apply Fraudar on the correlation matrix, treating it as a fully connected weighted graph. Although Fraudar does not target correlation, the algorithm keeps removing nodes of a graph with the least sum of edge weights plus node weights to search for a dense subgraph of high anomaly score, thus noticeable correlation could be expected if it can truly find the dense subgraph. We refer to the principal score of the anomaly set identified by Fraudar as *Fraudar PS*.

In addition, to deliberately show previous generative models for non-correlated group anomalies is ineffective for the correlated anomaly detection problem, we experiment with [3] on the first dataset since the dataset contains some rich attributes like request tokens, response, agent info, etc. We feed in the model with all such information. All such generative models do not have correlation in their consideration, and we cannot think of any part of its design could capture anomalous correlation, so we expect their failure.

For Randomized PS (rPS), we use $p = 1.4$ and 20% sampling rate; for Generative PS (gPS) we constrain $\frac{a}{a+b} \geq a = 0.75$ for the anomaly set. The alert threshold $\tilde{\rho}$ is set to 0.7 for all experiments, but we ignore alerts whose anomaly strength is less than 0.1% of the time window.

*A. Qualitative Results for Server Logs*

We run direct principal score evaluation and our randomized algorithms on the original data set with 19300+ 1-hr long sliding window and compare the detection results shown in Figure 5 to qualitatively confirm discussions in previous sections. The distinct IPs in each window can range from 300 to 4,000, and the average is 1260. We set the number of anomaly sets as $k = 2$ for this experiment. The PS by direct Lanczos algorithm raised 264 alerts (12 contains unique discovery), the randomized PS raised 5306 alerts (3205 "unique"), the Generative PS raised 2845 alerts (853 "unique"), and the Fraudar algorithm raised 1022 alerts (only 37 "unique"). The two major attacks, roughly between time windows 2500-3000 and 16000-17000 are well observed in all evaluations.

We give the following interpretations of the results. **1)** The pure recursive Lanczos method only responses to time windows where the correlated anomalies significantly differ from others (IPs that make much more requests than normal IPs). **2)** The random sampling technique introduced by rPS in Section III.C brings out the hidden anomalies, even if the IPs try to simulate human-like visit rates, as long as they prevail in the time window so that the "anomaly strength" defined in Section III.C is strong. **3)** both rPS and gPS discover much suspicious IPs that could be collecting real-time data from the server, but they might not the majority of the data, because the purpose is not to overload the server. In this case, as discussed earlier, the window principal score is overwhelmed by normal data, and therefore the Lanczos itself is not able to discover them, as discussed in section III.A and III.B. The generative model gPS makes conservative decisions, and is capable of a noticeable amount of unique discoveries. **4)** The Fraudar algorithm can capture correlation when applied on the IP-URL bipartite graph, even this is not considered in their design; however, its problem is many anomalous IPs in this dataset request the same or even less number of URLs than normal IPs in a window, so removing them can increase the anomaly score of the remaining, and thus many correlated anomalies are treated as normal by the algorithm; at last their detected anomaly set only contains those made most URL requests; **5)** Generative models like MGMM based on topic model with Gaussian mixture essentially does not model CAD and does not work on our problem.

*Table 1 Comparison of Average Recall, Estimated Accuracy and Runtime*

| Server Log Data Set | Big Anomaly Sets | | | Few Anomalies with High Strength | | | Hidden Anomalies | | | Overall | | |
|---|---|---|---|---|---|---|---|---|---|---|---|---|
| | Rec[2] | Acu[3] | Time[4] | Rec | Acu | Time | Rec | Acu | Time | Rec | Acu | Ex[6] |
| dPS[1] ($\tilde{\rho}=0.7$) | 847/0.79 | 0.835 | 45/236s | 458/0.38 | 0.735 | 30/145s | 33/0.04 | 0.506 | 26/124s | 0.45/0.65 | 0.78 | 76 |
| dPS ($\tilde{\rho}=0.6$) | 987/**0.93** | 0.704 | | 726/0.70 | 0.610 | | 86/0.06 | 0.442 | | 0.60/0.82 | 0.66 | 214 |
| Fraudar | 715/0.31 | 0.874 | **1.5/6.5s** | 862/0.87 | 0.853 | **1.7/7.9s** | 28/0.04 | 0.573 | **0.8/3.4s** | 0.54/0.45 | 0.86 | 130 |
| rPS ($\tilde{\rho}=0.7$) | 952/0.87 | 0.833 | 2.3/14s | 848/0.85 | 0.784 | 2.1/11s | 19/0.01 | 0.544 | **2.0/6.9s** | 0.62/0.83 | 0.81 | 85 |
| gPS | 801/0.64 | **0.940** | 11/19s | 603/0.56 | **0.915** | 8.7/14s | **724/0.68** | **0.807** | 4.1/7.5s | 0.71/0.61 | **0.88** | **13** |
| rPS + gPS[5] | **984**/0.90 | | | **924**/0.92 | | | 724/0.68 | | | **0.88**/0.90 | | |
| Stock Price Data Set | Big Anomaly Sets | | | Few Anomalies with High Strength | | | Hidden Anomalies | | | Overall | | |
| | Rec | Acu | Time | Rec | Acu | Time | Rec | Acu | Time | Rec | Acu | Ex |
| dPS[1] ($\tilde{\rho}=0.7$) | 325/0.53 | **0.852** | 56/320s | 97/0.55 | 0.733 | 49/252s | 28/0.04 | 0.527 | 35/145s | 0.48/0.69 | 0.82 | 20 |
| dPS ($\tilde{\rho}=0.6$) | 561/0.61 | 0.658 | | 123/0.70 | 0.651 | | 45/0.05 | 0.522 | | 0.58/0.74 | 0.75 | 54 |
| Fraudar | Zero detection on all cases. | | | | | | | | | | | |
| rPS ($\tilde{\rho}=0.7$) | 813/0.81 | 0.827 | **3.9/26s** | 884/0.84 | 0.814 | **4.1/20s** | 23/0.01 | 0.318 | **3.0/8.1s** | 0.61/0.83 | 0.80 | 23 |
| gPS | 706/0.68 | 0.846 | 14/26s | 782/0.72 | **0.928** | 13.2/22s | **815**/0.82 | **0.767** | 12.5/32s | 0.82/0.79 | **0.86** | **0** |
| rPS + gPS[5] | **840**/0.82 | | | **912**/0.89 | | | 820/0.82 | | | **0.89**/0.86 | | |

(1) "dPS" represents principal score evaluation by direct us of Lanczos algorithms by [20] and [19]; (2) number of raised alerts (max 1000) and % of detected simulated anomalies; (3) % of detected anomalies are from simulation (excluding suspicious anomalies detected from the real data); (4) the average runtime for detection over a time window and the max runtime; (5) merging the alerts and detected anomalies increases recall; (6) number of extra detected time windows from the "control group", an indicator of how likely a raised alert is a potential false alarm.

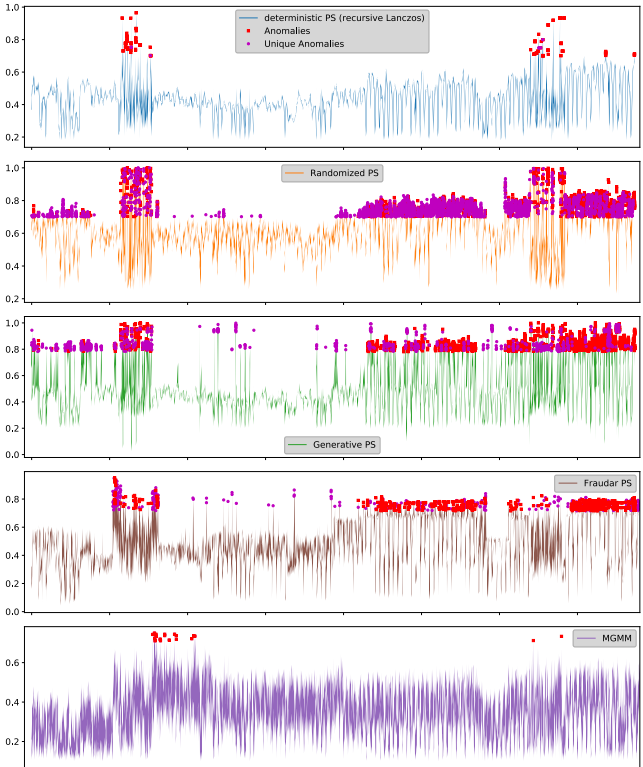

Figure 5 Correlated anomaly detection results from an ecommerce website server log data. Curves are principal scores of the anomaly set detected by each algorithm through the timeline. Red dots and magenta dots are alerts, where the later indicate the anomaly sets of the alert include unique anomalies detected by the algorithm.

We present the number of data entries and the number of detected anomalies of all alerts in Figure 6. For Lanczos PS and rPS, we consider the data entries whose correlation with the principal component higher than 0.7 as anomalies. For gPS, those clustered into the anomaly sets are considered as anomalies. It is clear from the chart in general gPS detects less anomalies than rPS at the time when they both raise alert, confirming the discussion of Figure 3.

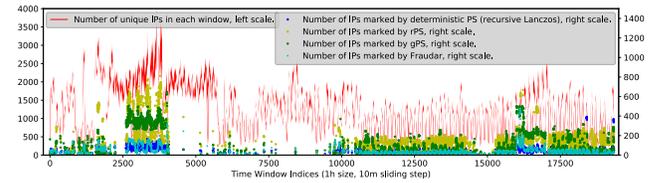

Figure 6 Number of data entries and the number of anomalies marked in each window. Note different scales are used for better visualization.

### B. Qualitative Results for Stock Price Data

The stock data set contains daily price series since 1970s. We use the data after 1983 so the number of stocks in each time window is more than 20. We define 30-business-day as a time window and let 3 business days as the sliding step, and use the closing price series. At the beginning, there are only dozens of stocks; new stocks appear over time, and at the end there are near 7000 stocks. This dataset has much lower background correlation than the logs, with average window-wide principal score around 0.28. We set $k=5$ for this experiment since we expect multiple stock sectors might exhibit correlation. The Fraudar algorithm can only be applied on the correlation matrix, treating it as a fully connected weighted graph. The results are similar to the server log data set. We can in addition observe the direct use of PS in Figure 7 makes no detection as the number of stocks in each window increase with time.

Besides, the Fraudar algorithm no longer works. We actually can claim Fraudar generally fails for any sufficiently densely connected graph, because the key line in their algorithm is to "remove a node that best increase the anomaly score of the remaining", which mathematically requires the anomaly score of the current subgraph -- the sum of edge weights divided by the number of nodes in it -- should be higher than the sum of the edge weights of the node to be removed, which is very unlikely for a dense graph. Therefore, the algorithm typically terminates prematurely, ending up with a large subgraph whose correlation is nearly the same as the background correlation.

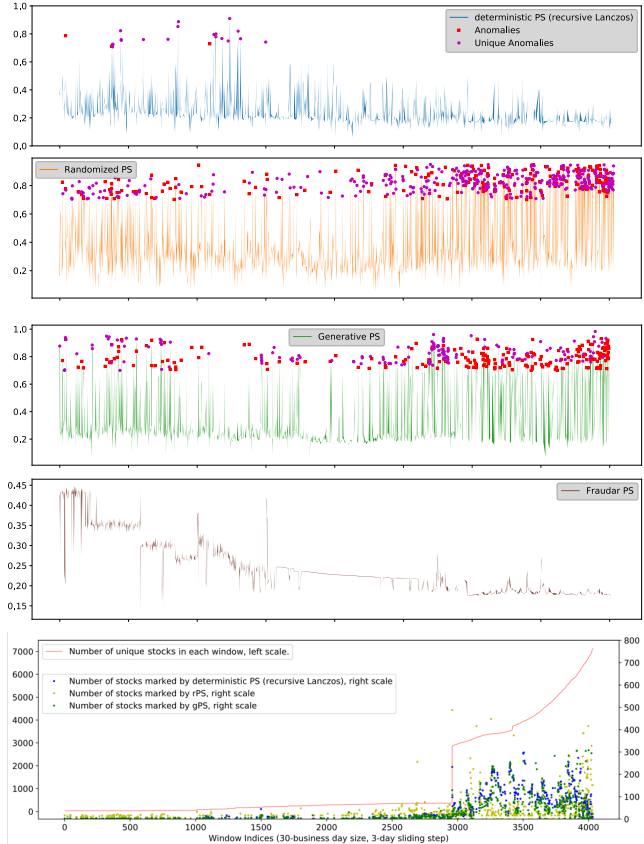

Figure 7 Correlated anomaly detection on a daily U.S. stock price data set, and the number of data entries/anomalies marked in each window.

*C. Quantitative Analysis*

To quantitatively evaluate performance, we have to inject simulated anomalies to the original data due to lack of gold standard and true labels in the data sets we have.

***For the server log data set***, we developed a software that truly visits the webpages hosted by the server, so the generated logs have realistic webpage request sequence and intervals. We collect the logs and inject them to 3000 randomly chosen 1-hr time windows in three cases, 1000 windows for each: **1)** in the "big anomaly sets" case, we inject anomalies of various anomaly strength to a window to take up 20% to 50% of the data entries; **2)** in the "strong anomaly strength" case, the anomalies take up 5% to 20% of the data with high anomaly strength, obeying Assumption 3; **3)** the "hidden anomalies" case injects 20~200 correlated anomalies with low anomaly strength. The three cases cover most of the situations for CAD on the log data. The parameter setup is the same as in the qualitative analysis. The "core" anomalies identified by gPS from real data are removed, while suspicious anomalies are kept for reasonable noise. We randomly sample additional 1000 windows without adding any simulation as the "***control group***".

***For the stock price data set***, we manually identify 100+ likely true anomaly sets (many of them consist of stocks from the same sector or related sectors), replicate them in other time windows with random scaling plus small random perturbations that affect but do not break the correlations, simulating three situations like what we do for server logs: 1) "big anomalies" mimics the situation when a large number of stocks fluctuate in a correlated way, which could happen when the market suffers a big turbulence; 2) "strong anomaly strength" is the case when a few stocks simultaneously have big price change; 3) "hidden anomalies" is the case when a few stocks have mild to tiny correlated price changes. Again 1000 30-business-day time windows are sampled for each case for injection, with 1000 additional windows as the "control group" without injection. Anomaly sets from the original data identified by gPS are removed for the sampled windows.

Comprehensive ***performance comparison*** results are shown in Table 1.
1) We measure how many alerts are raised for each method, how many simulated anomalies can be recognized.
2) We estimate the concept of "***accuracy***" by calculating the percentage of the detected anomalies are from the simulation excluding suspicious data entries detected from the real data; after such exclusion, remaining real data entries are very likely to be non-anomalies.
3) We count how many alerts are raised for the "control group", which is an indicator how likely each method could give false alarms.
4) Average runtime is also recorded since time cost is an important consideration for real-time analysis.

The evaluation result shows our framework attains best performance in terms of recall, the estimated "accuracy" and runtime. We have a few remarks for the results:
1) It confirms naïvely lowering alert threshold would not be a good idea as it introduces noticeable false alarms;
2) rPS is competitive with PS for big anomaly sets, and excels in the case of assumption 3; gPS is confirmed to be a more conservative method, but its high accuracy makes its alerts and detected anomalies most trustable; in the hidden case, only gPS is properly working; as a result, our detection framework rPS+gPS makes a great partner.
3) The time cost is dramatically reduced, which makes it possible to use the same computational resource to monitor much more data;

***The scaling capability*** of rPS and gPS is shown in Figure 8. We randomly sample 100 windows with simulated anomalies recognizable by all methods (such window mostly falls in the first case of Table 1), and scale the window size up to three times by adding likely non-anomalous data. From the left chart, the direct PS quickly lose its effectiveness as the data size grows, while rPS and gPS maintains reasonable performance, a result consistent with our discussion in section III.B and Figure 2. For time use, both rPS and gPS significantly outperforms direct PS. Theoretically, the direct PS at best has sub-cubic time complexity [12, 19, 27]; rPS is sub-cubic time complexity of a small coefficient due to the sampling, and gPS is of quadratic time of small coefficient proved in Proposition 3. Therefore, the scalability and time saving of our framework is a definite advantage for large-scale streaming data processing.

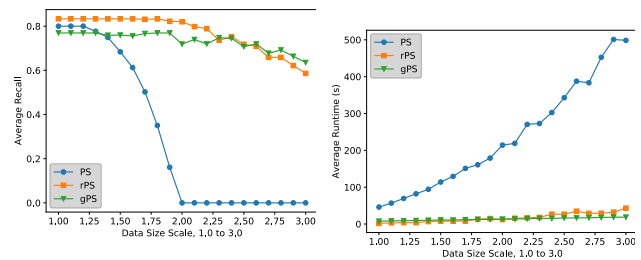

Figure 8 Scalability of direct PS, rPS, and gPS. Left: trends of recall as data size grows. Right: time use as data size grows.

*As last, we present parameter tuning* of our framework in Table 2, including the sampling rate $r$, the norm order $p$ for anomaly strength calculation, the constraint $a$ on beta mean, and the number of anomaly set $k$. The choice of the best parameter is the one with best balance between recall and accuracy. The best $r$ can be viewed as a reflection of percentage of anomalies. A larger $p$ can help recover more anomalies, but in the meantime reduces accuracy. The $a$ is the average correlation for two data entries in an anomaly set, whose best value is no surprisingly near the near the alert threshold $\tilde{p}$. Last, the result for $k$ confirms generally there should not be many anomaly sets in one time window.

| Table 2 Parameter Tuning Results | | | | | | | |
|---|---|---|---|---|---|---|---|
| $r$ | Rec | Acu | Ex | $p$ | Rec | Acu | Ex |
| 0.05 | 0.722 | 0.822 | 78 | 1 | 0.632 | 0.801 | 45 |
| 0.1 | 0.803 | 0.801 | 82 | 1.2 | 0.770 | 0.788 | 72 |
| **0.2**[1] | **0.851** | **0.784** | **85** | **1.4** | **0.853** | **0.784** | **85** |
| 0.3 | 0.814 | 0.752 | 85 | 1.6 | 0.884 | 0.713 | 124 |
| 0.4 | 0.766 | 0.747 | 83 | 1.8 | 0.915 | 0.602 | 185 |
| 0.5 | 0.705 | 0.741 | 80 | 2 | 0.943 | 0.541 | 275 |
| $a$ | Rec | Acu | Ex | $k$ | Rec | Acu | Ex |
| 0.6 | 0.727 | 0.787 | 51 | 1 | 0.56 | 0.915 | 13 |
| 0.65 | 0.672 | 0.822 | 35 | **2** | **0.53** | **0.924** | **11** |
| 0.7 | 0.619 | 0.874 | 26 | 3 | 0.512 | 0.939 | 10 |
| **0.75** | **0.568** | **0.915** | **13** | 4 | 0.488 | 0.957 | 10 |
| 0.8 | 0.524 | 0.923 | 11 | 5 | 0.425 | 0.966 | 10 |
| 0.85 | 0.490 | 0.935 | 11 | 6 | 0.35 | 0.97 | 10 |
| (1) The chosen parameters for our previous experiments are highlighted. | | | | | | | |

## V. CONCLUSION & FUTURE WORK

We target correlated anomaly detection (CAD) in this paper. We realize previous approaches cannot be trivially applied on large-scale streaming data. The paper first brings up the issue of principal score degeneration and contributes a rigorous discussion. We then use what is discovered from the proves to develop two randomized algorithms rPS and gPS. The experiments on two large data sets verify our methods perform much better than direct calculation of principal score, and some other recent group anomaly detection algorithms. Besides performance boost, rPS and gPS significantly reduce the computation time and thus also more scalable to large data.

Our future work can prove the convergence of gPS algorithm. This is a mathematically interesting model. When the label vector **z** is fixed, it is a fixed-point iteration, and we can prove the convergence by showing the fixed-point iteration formulas are continuous differentiable and further a contraction. However, when **z** is not fixed, the problem is more difficult. Another potential future work is anomalous subgraph detection on a densely weighted network, as we have seen it can model or even equate the CAD problem. We have shown the recent model Fraudar fails for this problem, and our model gPS can serve as the basis for the research.

**Acknowledgement**: This work was supported in part by NSF III 1815256, NSF III 1744661, NSF CNS 1650431.